\documentclass[letterpaper, 10 pt, conference]{ieeeconf}
\IEEEoverridecommandlockouts       

\usepackage[usenames,dvipsnames,svgnames]{xcolor}
\usepackage{times} 
\usepackage{amsmath} 
\usepackage{amssymb}  
\usepackage{algorithm}
\usepackage[noend]{algpseudocode}
\usepackage{comment}
\usepackage{graphicx}
\usepackage{subfigure}
\usepackage{tikz}
\usepackage{etoolbox}
\usepackage{xspace}
\usepackage[letterspace=0,tracking=true]{microtype}
\usepackage{comment}
\usepackage{hyperref}
\usepackage[utf8]{inputenc}
\usepackage[T2A,T1]{fontenc}
\usepackage{booktabs}
\usepackage{paralist}
\usepackage[capitalise]{cleveref}
\usepackage[export]{adjustbox}
\usepackage{array}
\usepackage{pgfplots}

\makeatletter
\let\NAT@parse\undefined
\makeatother
\usepackage[compress,numbers]{natbib}

\hypersetup{%
  pdfborder = {0 0 0},
  bookmarksdepth = section,
  citecolor = DarkGreen,
  linkcolor = DarkBlue,
  urlcolor = DarkBlue,
}%

\graphicspath{{images/}}

\definecolor{Red}{rgb}{1,0,0}
\definecolor{Green}{rgb}{0,0.69,0}
\definecolor{Blue}{rgb}{0,0,1}
\definecolor{LightBlue}{rgb}{0,0.5,1}
\definecolor{veryLightBlue}{rgb}{0.85,0.98,1}
\definecolor{veryLightGreen}{rgb}{0.6,1,0.6}
\definecolor{Skin}{rgb}{1,0.71,0.69}
\definecolor{Grey}{rgb}{0.5,0.5,0.5}
\definecolor{LightGrey}{rgb}{0.6,0.6,0.6}
\definecolor{Black}{rgb}{0,0,0}
\definecolor{White}{rgb}{1,1,1}
\newcommand{\red}{\color{Red}}

\newbool{blind}
\setbool{blind}{false}

\ifbool{blind}
{\AtBeginEnvironment{blind}{\comment}%
  \AtEndEnvironment{blind}{\endcomment}}{}

\makeatletter
\DeclareRobustCommand\onedot{\futurelet\@let@token\@onedot}
\def\@onedot{\ifx\@let@token.\else.\null\fi\xspace}

\newcommand{\eg}{e.g.,\xspace}

\makeatother

\newcommand{\TODO}[1]{{\red TODO: {#1}\xspace }}

\title{\LARGE \bf
  Trajectory Prediction with Linguistic Representations
}

\author{Yen-Ling Kuo$^{*,1,2}$, Xin Huang$^2$, Andrei Barbu$^2$, Stephen G. McGill$^1$\\Boris Katz$^2$, John J. Leonard$^{1,2}$, Guy Rosman$^1$%
  \thanks{$^1$Toyota Research Institute, Cambridge, MA 02139, USA}%
  \thanks{\hspace*{2ex}\texttt{\{guy.rosman,stephen.mcgill,john.leonard\}@tri.global}}%
  \thanks{$^2$CSAIL \& CBMM MIT, Cambridge, MA 02139, USA}%
  \thanks{\hspace*{2ex}\texttt{\{ylkuo,huangxin,abarbu,boris\}@mit.edu}}%
  \thanks{$^*$Work done as an intern at TRI.}%
}%

\usetikzlibrary{calc}
\usetikzlibrary{fit}
\usetikzlibrary{positioning}
\usetikzlibrary{decorations.markings}
\usetikzlibrary{shapes.geometric}
\usetikzlibrary{shapes.multipart}
\usetikzlibrary{bayesnet}
\usetikzlibrary{tikzmark}
\usetikzlibrary{backgrounds} 
\usetikzlibrary{shapes}
\usetikzlibrary{3d}
\usetikzlibrary{arrows.meta}
\usetikzlibrary{shapes.geometric}

\definecolor{pinegreen}{cmyk}{0.92,0,0.59,0.25}
\definecolor{royalblue}{cmyk}{1,0.50,0,0}

\definecolor{Red}{rgb}{1,0,0}
\definecolor{Green}{rgb}{0,0.69,0}
\definecolor{Blue}{rgb}{0,0,1}
\definecolor{LightBlue}{rgb}{0,0.5,1}
\definecolor{veryLightBlue}{rgb}{0.85,0.98,1}
\definecolor{veryLightGreen}{rgb}{0.6,1,0.6}
\definecolor{Skin}{rgb}{1,0.71,0.69}
\definecolor{Grey}{rgb}{0.5,0.5,0.5}
\definecolor{LightGrey}{rgb}{0.6,0.6,0.6}
\definecolor{Black}{rgb}{0,0,0}
\definecolor{White}{rgb}{1,1,1}

\newif\ifcomments
\commentstrue
\ifcomments
        \newcommand{\GR}[1]{\color{cyan}(GR: #1)\color{black}\xspace}
        \newcommand{\YL}[1]{\color{purple}(YLK: #1)\color{black}\xspace}
        \newcommand{\AB}[1]{\color{olive}(AB: #1)\color{black}\xspace}
        \newcommand{\JL}[1]{\color{magenta}(JL: #1)\color{black}\xspace}
        \newcommand{\SM}[1]{\color{blue}(SM: #1)\color{black}\xspace}
        \newcommand{\CH}[1]{\color{orange}(CH: #1)\color{black}\xspace}
        \newcommand{\BK}[1]{\color{violet}(CH: #1)\color{black}\xspace}
\else
        \newcommand{\GR}[1]{}
        \newcommand{\YL}[1]{}
        \newcommand{\AB}[1]{}
        \newcommand{\JL}[1]{}
        \newcommand{\SM}[1]{}
        \newcommand{\CH}[1]{}
        \newcommand{\BK}[1]{}
\fi

\newcommand{\rev}[1]{{\color{black} #1}}

\begin{document}

\maketitle
\thispagestyle{empty}
\pagestyle{empty}

\begin{abstract}
  Language allows humans to build mental models that interpret what is happening
  around them resulting in more accurate long-term predictions.
  We present a novel trajectory prediction model that uses linguistic
  intermediate representations to forecast trajectories, and is trained using
  trajectory samples with partially-annotated captions.
  The model learns the meaning of each of the words without direct per-word
  supervision.
  At inference time, it generates a linguistic description of trajectories which
  captures maneuvers and interactions over an extended time interval.
  This generated description is used to refine predictions of the trajectories
  of multiple agents.
  We train and validate our model on the Argoverse dataset, and demonstrate
  improved accuracy results in trajectory prediction.
  In addition, our model is more interpretable: it presents part of its
  reasoning in plain language as captions, which can aid model development and
  can aid in building confidence in the model before deploying it.
\end{abstract}

\section{INTRODUCTION}

Predicting future trajectories of agents on the road is indispensable for autonomous driving, allowing a vehicle to plan safe and effective actions.
Prediction requires understanding of how agents relate to a variety of concepts, such as map features, other agents, maneuvers, rules, to name a few.

Language allows us to reason about such concepts easily. Humans can use language to richly describe events in the world.
These descriptions can involve other agents, actions, goals, and objects in the
environment.
They also focus our attention on entities, relationships, and properties,
cutting through a sea of irrelevant details.
Finally, language is also flexible in its treatment of time, as it can
seamlessly describe the past and multiple possible futures within a single
utterance.
%

%

%
%
%
%

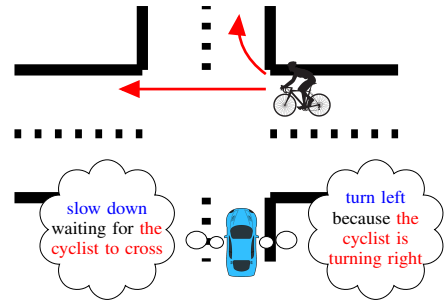
\begin{figure}[htp]
  \centering
  \scalebox{0.85}{%
    \begin{tikzpicture}
      \draw[line width=5] (0,0) -- (2,0);
      \draw[line width=5] (4,0) -- (6,0);
      \draw[line width=4, loosely dashed] (0,1) -- (2,1);
      \draw[line width=5] (0,2) -- (2,2);
      \draw[line width=5] (4,2) -- (6,2);
      \draw[line width=4, loosely dashed] (4,1) -- (6,1);
      \draw[line width=5] (2,2) -- (2,3);
      \draw[line width=5] (4,2) -- (4,3);
      \draw[line width=4, loosely dashed] (3,2) -- (3,3);
      \draw[line width=5] (2,0) -- (2,-1);
      \draw[line width=5] (4,0) -- (4,-1);
      \draw[line width=4, loosely dashed] (3,0) -- (3,-1);
      \node (car) at (3.55, -0.7) {
        \includegraphics[width=.03\textwidth]{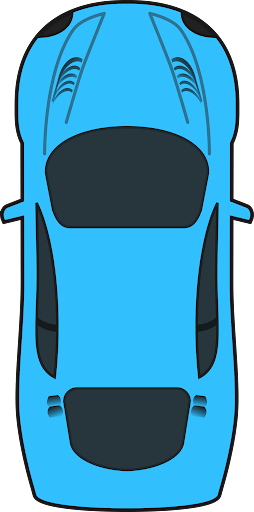}
      };
      \node (cyclist) at (4.5, 1.7) {
        \includegraphics[width=.05\textwidth]{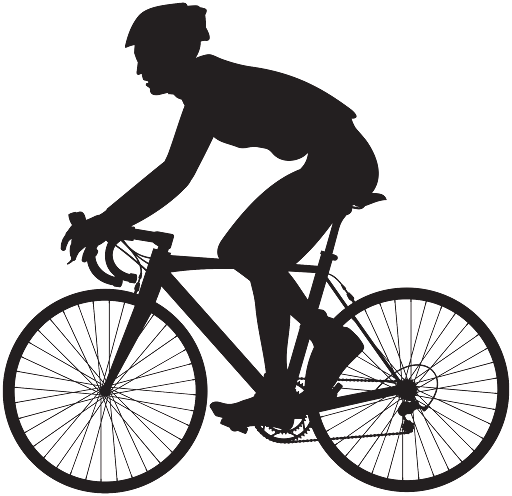}
      };
      \node (end1) at (1.5, 1.7) {};
      \node (end2) at (3.5, 3) {};
      \draw[->, red, very thick] (cyclist) -- (end1);
      \draw[->, red, very thick] (cyclist) to [bend left] (end2);
      \node[cloud callout,draw,inner sep=-8pt, fill=white,callout absolute pointer=(car.east), font=\footnotesize,text width=53pt,align=center] (car-callout1) at ($(car)+(2.1, 0.2)$) { \textcolor{blue}{turn left} because \textcolor{red}{the cyclist is turning right} };
      \node[cloud callout,draw,inner sep=-8pt, fill=white,callout absolute pointer=(car.west),font=\footnotesize,text width=53pt,align=center] (car-callout1) at ($(car)-(2.1, -0.2)$) { \textcolor{blue}{slow down} waiting for \textcolor{red}{the cyclist to cross} };
    \end{tikzpicture}
  }
  \caption{\rev{An illustration of the idea that} an ego-car can use language to hypothesize possible futures by applying different affordances of language. The words in \textcolor{blue}{blue} shows the ego-car's reasoning about \textcolor{blue}{its own actions} conditioned on the \textcolor{red}{other agent's behavior}, shown in \textcolor{red}{red}. The words such as ``waiting for'' and ``because'' allow the ego-car to reason about the temporal relationship and causal relationship in this interacting scenario. }
  \label{fig:language_scenario}
\end{figure}

Thus far, models for reasoning about road scenarios have not availed themselves
of the added level of abstraction that language provides.
A concrete example where language can help is shown in
\cref{fig:language_scenario}.
The ego-vehicle is approaching an intersection while a cyclist is about to pass
through the intersection.
The car must predict the future trajectory of the cyclist and its own movements
in response to that trajectory.
\rev{The hypothetical futures of the ego-vehicle can be described in} sentences such as ``I should slow
down waiting for the cyclist to cross and then turn left'' or ``I can turn left
because the cyclist is turning right''.
%
%
%
Having access to the descriptions of two scenarios enables one to quickly
check which of them is happening by removing many irrelevant details.
This is the power that language brings to trajectory \rev{prediction}, it abstracts away
irrelevant non-actionable information, and focuses the computation on relevant
parts such as the temporal order of crossings and interactions with other road
agents.
%
%
%

Taking the notion of language as an abstraction for agents, interactions, and temporal reasoning, we propose a predictor, illustrated in \cref{fig:overall-model}, which leverages a learned linguistic representation to attend to the relevant agent and the relevant parts of descriptions at each time step.
We co-train the sentence generator with the attention-based trajectory predictor to generate sentences that hypothesize future trajectories.
Every generated word is grounded in the future trajectories of the
agents.
\rev{We train and test the proposed predictor on the Argoverse dataset with a set of synthetic linguistic tokens.}
Inspecting the attention on each word reveals 1) what concepts are relevant for
the current predictions, and 2) what properties or
interactions those concepts are based on.
This greatly accelerates model building: rather than dealing with the failure of
a black box one can discover that a particular word is not well trained.
It also provides an auditing mechanism, which will help us to ensure that autonomous
cars are making the right decisions for the right reasons, rather than
incidentally succeeding on benchmarks, thus risking failure in the real world.
Finally, it provides an audit that traces the reasoning of the model which can
be helpful when understanding accidents.
%
%

%
%
%

This work makes the following contributions.
\begin{enumerate}
  \item We develop trajectory-prediction datasets augmented with linguistic descriptions.
  \item We present a novel predictor that leverages language as one of its internal representations.
  \item We demonstrate how a spatio-temporal attention on top of that language can help trajectory prediction.
\end{enumerate}



\section{RELATED WORK}
\label{sec:related-work}

\subsection{Trajectory Prediction with Attention and Structures}

Trajectory prediction of road agents has become an important research topic in recent years.
Different inputs to trajectory prediction models such as providing past trajectories \cite{scholler2020constant}, maps \cite{gilitschenski2020deep,gao2020vectornet,liang2020learning}, and images \cite{bojarski2016end,amini2019variational,huang2019uncertainty} have been considered.
Structural priors that encapsulate temporal and inter-agent reasoning have been incorporated, implicitly with attention \cite{alahi2016social,gupta2018social,sadeghian2019sophie,amirian2019social,mercat2020multi} and explicitly via maneuvers and rules \cite{deo2018multi,huang2020diversitygan,li2021vehicle, huang2019uncertainty, tang2019multiple, salzmann2020trajectron++,huang2021hyper} and goals \cite{rhinehart2019precog,zhao2020tnt}.
%
Language is a structural prior, but it is one that spans all of these topics
(naturally and flexibly combining interactions, maneuvers, and goals) and
integrates seamlessly with attention.
This is not just for the ease of communication but also allows us to express all possible occurrences in the physical world without being confined to only object references or any specific temporal structure.

Similar to attention-based predictors~\cite{li2021rain,mercat2020multi}, attention plays a key role in our model and guides the trajectory predictions.
Although the scope of the attention module is gated by the linguistic
descriptions~\cite{mun2017text,wang2019neighbourhood,kuo2020deep}, agents that don't co-occur in the same description don't need to
be jointly attended to.
Language guides attention which ultimately removes extraneous agents
and improves performance.
In addition, the pooled agent states are combined with the word embeddings of
descriptions to generate attention over the relevant language tokens that
represents sequences of events.
Our language-guided attention approach efficiently combines both social and
temporal reasoning for trajectories while prior work only considers one of these
at a time.
%


\subsection{Language and Planning}

Most prior work that uses language in robotics involves planning to follow natural language commands in various tasks including manipulation~\cite{tellex2011understanding,paul2017temporal,paxton2019prospection,kuo2020deep,Stepputtis:goalmap} and navigation~\cite{tellex2011understanding,blukis2018mapping,magassouba2021crossmap}.
These planners mainly learn from demonstrations paired with input linguistic commands, i.e. learning an action distribution conditioned on linguistic input.
Our approach explores another language use, where language is implicit not explicit --- no linguistic input is provided; instead language acts as an internal representation for hypotheses about future trajectories.
To that end, our approach includes a sentence generator while these planners do not.
There are approaches that generate captions for videos~\cite{barbu2012video,yu2016video,chen2019temporal} but they have not yet been considered in planning.
The input linguistic commands for planners mostly specify goals and referents, e.g., landmarks, for navigation or locating the objects to interact with.
Some planners learn visual attention maps conditioned on language to predict a goal map~\cite{Stepputtis:goalmap}, filter relevant objects~\cite{kuo2020deep}, or estimate a visitation map~\cite{blukis2018mapping} for a single agent.
However, none of these planners considers language describing interactions with other agents and manners of performing an action, such as quickly or slowly.
These linguistic attributes are critical for trajectory predictions.
In addition to grounding words to visual maps, our model grounds words to agents' own trajectories and interactions with others.

\subsection{Neuro-symbolic Networks \& Architectures}
Our approach relates to different uses of neuro-symbolic architectures in other fields such as analysis of images~\cite{karpathy2015deep,hendricks2016deep,amizadeh20a} and videos~\cite{kulal2021hierarchical,Yi2020CLEVRER}.
Recent work by \citet{chitnis2021learning} uses symbolic planning to guide the
continuous planning process.
These architectures combine neural components with symbolic representations such as programs or domain specific languages.
Our approach similarly uses a symbolic component to guide the continuous component, i.e. trajectory prediction, but our approach works not only for a domain specific language but also for natural languages.

\section{MODEL}

\begin{figure*}[htp]
    \centering
    \scalebox{0.75}{
    \begin{tikzpicture}
      \node (agent) {
        \includegraphics[width=0.02\textwidth]{images/blue-car.png}
      };
      \node[below=0.3 of agent] (map) {
        \includegraphics[width=0.08\textwidth]{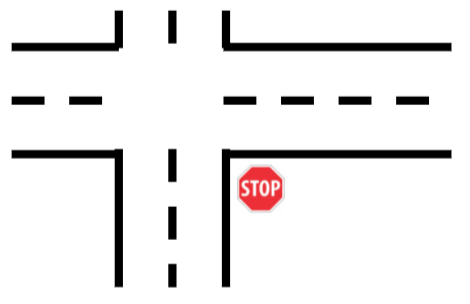}
      };
      \node[rounded corners, inner sep=5pt, text width=40, align=center, fill=cyan, text=white, font=\sffamily, right=1.1 of agent] (agent-encoder) {Agent Encoder};
      \node[rounded corners, inner sep=5pt, text width=40, align=center, fill={rgb:red,0.0;green,0.85;blue,0.34}, text=white, font=\sffamily, right=0.6 of map] (map-encoder) {Map Encoder};
      \node[,rounded corners, inner sep=5pt, text width=40, align=center, fill=red!50, text=white, font=\sffamily] (gnn-encoder) at ($(agent-encoder)+(2.5,-0.7)$) {GNN Encoder};
      \node[align=center] (encoder-text) at ($(gnn-encoder)+(-0.8,1.9)$) {Encoder};
      \node[draw, minimum height=40, fill=red!20, right=0.6 of gnn-encoder] (h0) {};
      \node[align=center, above=0.05 of h0] (h0-text) {$h_0$};
      \node[rounded corners, inner sep=5pt, text width=45, fill=magenta!60, text=white, font=\sffamily, align=center, right=0.7 of h0] (token-generator) {Sentence Generator};
      \node[below=0.5 of token-generator] (noise-token) {noise};
      \node[text width=50, right=1.7 of token-generator] (tokens) {\emph{Slow down, yield to Agent\#1, turn left.}};
      \node[align=center, above=0.05 of tokens] (tokens-text) {Tokens};
      \node[rounded corners, inner sep=5pt, text width=42, fill=orange!70, text=white, font=\sffamily, align=center, right=0.7 of tokens] (traj-decoder) {Trajectory Decoder};
      \node[below=0.5 of traj-decoder] (noise-traj) {noise};
      \node[right=0.7 of traj-decoder] (traj) {$\mathbf{s}_F$};

      \begin{scope}[on background layer]
        \node (encoder) [rounded corners, dashed, draw=black, fit= (agent-encoder)(map-encoder)(gnn-encoder)(h0), inner sep=2ex] {};
      \end{scope}

      \draw[->] (agent) -- (agent-encoder);
      \draw[->] (map) -- (map-encoder);
      \draw[->] (agent-encoder) -- (gnn-encoder);
      \draw[->] (map-encoder) -- (gnn-encoder);
      \draw[->] (gnn-encoder) -- (h0);
      \draw[->] (h0) -- (token-generator);
      \draw[->] (noise-token) -- (token-generator);
      \draw[->] (token-generator) -- node[midway,above] {sample} (tokens);
      \draw[->] (tokens) -- (traj-decoder);
      \draw[->] (noise-traj) -- (traj-decoder);
      \path (traj-decoder) edge [loop above] node {$h_t$} (traj-decoder);
      \draw[->] (traj-decoder) -- (traj);

      \draw[rounded corners=5pt,inner sep=10ex,-{Latex[length=1.2ex,width=1.2ex]}] (h0.south) -- node[] {} ++(0,-0.75) -| ($(traj-decoder.south)+(-0.5,0)$);
    \end{tikzpicture}}
  \vspace*{-7ex}
    \caption{An overview of the model with language as its intermediate representation. The encoder consists of a model which embeds the observed trajectory of each agent along with the map into $h_0$. A sentence is predicted from $h_0$. Together, the sentence and $h_0$ are used to refine the trajectory using the trajectory decoder; details of which are provided in \cref{fig:attention-module}. The result is a predicted trajectory $\mathbf{s}_F$. \rev{We show only one agent in this figure but all agents (up to four agents in our experiment) in the scene are encoded using the same encoder.}}
    \label{fig:overall-model}
\end{figure*}

\subsection{Problem Formulation}
\paragraph{Trajectory Prediction}
Given a sequence of the observed states for an agent
$\mathbf{s}_P = \{s_{T^{\prime}+1}, \cdots, s_0\}$, our goal is to predict $N$
possible future trajectories, $\mathbf{s}_F$.
Each agent trajectory is a sequence, $\mathbf{s}_F = \{s_1, \cdots, s_T\}$, where
$t$ is time and $s_t$ denotes the state of an agent at time $t$.
$t=0$ corresponds to the last observed time, i.e. the time at which we make the
predictions about the future.
The agent's past interactions with other agents and map elements are encoded
into an embedding $h_0$ as shown in \cref{fig:overall-model}.

\paragraph{Learning Intermediate Representation}
In addition to predicting future trajectories given the observed states $\mathbf{s}_P$, we aim to learn to sample $N$ possible linguistic descriptions $Y_{1:M}$ that matches how people describe the agent's future rollouts and use the descriptions to predict trajectories $\mathbf{s}_F$.
This linguistic description is a sequence of tokens that describe the potential future states of the agent,  $Y_{1:M} = (y_1, \cdots, y_M), y_m \in \mathcal{Y}$, where $\mathcal{Y}$ is the vocabulary of candidate tokens to describe a driving scenario and $M$ is the maximum number of tokens in the description.

This description about the future is often generated recursively from the past context~\cite{sutskever2014sequence} as:
\begin{equation}
    \log p(Y_{1:M}|\mathbf{s}_P) = \sum_{m=1}^M \log p(y_m|y_{1:m-1}, h_0)
\end{equation}
While it is possible to generate exhaustive descriptions for every detail in every scene, we consider descriptions from a single agent's point of view with a length cap that ensures that only the most relevant concepts will be included.
As each generated description describes a hypothetical future, we can generate a sample of future trajectory $\mathbf{s}_F^n$ based on each description: $p(\mathbf{s}_F^n|Y_{1:M}^n)$.

\subsection{Model Overview}
The overall network structure is shown in \cref{fig:overall-model}.
Our model is based on a multi-agent encoder-decoder GAN~\cite{gupta2018social}, with an LSTM for each agent.
The generator uses an encoder-decoder structure to produce samples of future trajectories.
For the latent space representation in the encoder-decoder, in addition to using a regular fixed-dimension tensor output by the encoder, we generate linguistic descriptions to represent the future trajectory which is ultimately decoded by the trajectory decoder.
The encoder uses a graph neural network to encode inter-agent interactions and agents' relation to the map.
We use a joint random noise vector in sentence generation and the decoder to enable joint probabilistic modeling.
The discriminator determines whether the generated trajectory is realistic or not.

\subsection{Encoder}
We encode map information \rev{with attention over the road graph}~\cite{gao2020vectornet}.
This map encoder takes a map input as a set of lane centerlines and uses self-attention to combine the encoded tensors from all lane centerlines.
\rev{The map encodings are concatenated with the position encoding at each time point and then passed through an LSTM} \cite{alahi2016social} to get the hidden state vector $h_0$ at the last observed time $t=0$.
This hidden vector encapsulates the agent's observations up until $t=0$ and is used to generate linguistic descriptions and future trajectories.

\subsection{Language Generation}
The sentence generator is an LSTM.
It takes each agent's encoded hidden state vector $h_0$ along with a noise vector as input, to make $M$-step rollouts to generate a description of length $M$.
When there are multiple sentences in the description, we use two special tokens <bos> and <eos> to indicate the beginning and the end of a sentence and <pad> to pad the sequence if the description has less than $M$ tokens.
At each rollout step, the LSTM outputs a distribution over the vocabulary and we leverage Gumbel-softmax~\cite{jang2016categorical} to sample a token from the predicted distribution.

\subsection{Language Conditioned Decoder}
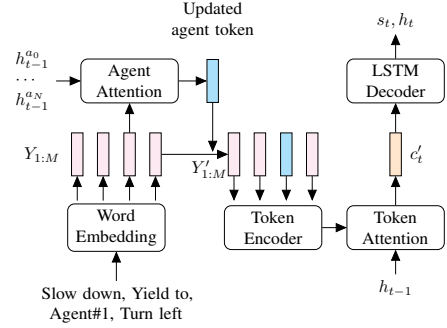
\begin{figure}[htp]
    \centering
    \scalebox{0.65}{
    \begin{tikzpicture}
      \node[align=center, text width=100] (tokens) {Slow down, Yield to, Agent\#1, Turn left};
      \node[draw, rounded corners, align=center, text width=50, above=0.6 of tokens] (word-embedding) {Word Embedding};
      \node[draw, minimum height=25, fill=magenta!10] (y0) at ($(word-embedding)+(-0.8,1.5)$) {};
      \node[draw, minimum height=25, fill=magenta!10] (y1) at ($(word-embedding)+(-0.27,1.5)$) {};
      \node[draw, minimum height=25, fill=magenta!10] (y2) at ($(word-embedding)+(0.27,1.5)$) {};
      \node[draw, minimum height=25, fill=magenta!10] (y3) at ($(word-embedding)+(0.8,1.5)$) {};
      \node[left=0.1 of y0] {$Y_{1:M}$};
      \node[draw, rounded corners, align=center, text width=50, above=0.6 of y2] (agent-attention) {Agent Attention};
      \node[text width=20, left=0.5 of agent-attention] (hs) {$h_{t-1}^{a_0}$\\$\cdots$\\$h_{t-1}^{a_N}$};
      \node[draw, minimum height=25, fill=cyan!30] (y2-agent) at ($(agent-attention)+(1.7,0)$) {};
      \node[text width=50, align=center, above=0.3 of y2-agent] {Updated agent token};
      \node[draw, rounded corners, align=center, text width=50, right=1.2 of word-embedding] (token-lstm) {Token Encoder};
      \node[draw, minimum height=25, fill=magenta!10] (y0-new) at ($(token-lstm)+(-0.8,1.5)$) {};
      \node[draw, minimum height=25, fill=magenta!10] (y1-new) at ($(token-lstm)+(-0.27,1.5)$) {};
      \node[draw, minimum height=25, fill=cyan!30] (y2-new) at ($(token-lstm)+(0.27,1.5)$) {};
      \node[draw, minimum height=25, fill=magenta!10] (y3-new) at ($(token-lstm)+(0.8,1.5)$) {};
      \node (new-ys) at ($(y0-new)-(0.5,0.3)$) {$Y_{1:M}^\prime$};
      \node[draw, rounded corners, align=center, text width=50, right=0.5 of token-lstm] (token-attention) {Token Attention};
      \node[below=0.5 of token-attention] (ht) {$h_{t-1}$};
      \node[draw, minimum height=25, fill=orange!20] (context) at ($(token-attention)+(0,1.5)$) {};
      \node[right=0.05 of context] (context-text) {$c_t^{\prime}$};
      \node[draw, rounded corners, align=center, text width=50, above=0.6 of context] (traj-lstm) {LSTM Decoder};
      \node[above=0.5 of traj-lstm] (st) {$s_{t}, h_t$};

      \draw[->] (tokens) -- (word-embedding);
      \draw[->] ($(y0.south)-(0,0.5)$) -- (y0);
      \draw[->] ($(y1.south)-(0,0.5)$) -- (y1);
      \draw[->] ($(y2.south)-(0,0.5)$) -- (y2);
      \draw[->] ($(y3.south)-(0,0.5)$) -- (y3);
      \draw[->] (y2) -- (agent-attention);
      \draw[->] (hs) -- (agent-attention);
      \draw[->] (agent-attention) -- (y2-agent);
      \draw[->] (y3) -- (y0-new);
      \draw[->] (y2-agent) -- ($(y2-agent.south)-(0,1)$);
      \draw[->] (y0-new) -- ($(y0-new.south)-(0,0.5)$);
      \draw[->] (y1-new) -- ($(y1-new.south)-(0,0.5)$);
      \draw[->] (y2-new) -- ($(y2-new.south)-(0,0.5)$);
      \draw[->] (y3-new) -- ($(y3-new.south)-(0,0.5)$);
      \draw[->] (token-lstm) -- (token-attention);
      \draw[->] (ht) -- (token-attention);
      \draw[->] (token-attention) -- (context);
      \draw[->] (context) -- (traj-lstm);
      \draw[->] (traj-lstm) -- (st);
    \end{tikzpicture}}
    \vspace{-2ex}
    \caption{Attention computation inside the trajectory decoder block in Fig.~\ref{fig:overall-model}. We use the tokens that refer to other agents to pool the agents' states first and then employ attention over tokens to determine the tokens that are relevant to the current time point to make a prediction.}
    \label{fig:attention-module}
\end{figure}

The decoder for each agent takes the language token sequence and embeds each of the tokens with an embedding layer.
%
%
%
We co-train the word embedding with the decoder.
The sentence description contains words that refer to other agents and the structure to relate the dependency of events.
We design the attention mechanisms to capture these language properties as illustrated in Fig.~\ref{fig:attention-module}.

\paragraph{Agent Attention}
When generating descriptions, we refer to an agent by its index number in the scenario.
Inside the decoder, when we see a word embedding $y_m$ that corresponds to an agent index $a$, we use an MLP layer with softmax output to generate attention over the agent to pool the relevant agent states in the previous time step $t-1$ to update the word embedding:
\begin{equation}
    y_m^\prime =  \textit{Softmax}(\textit{MLP}(y_m)) \cdot h_{t-1}^\mathcal{A}
\end{equation}
where $h_{t-1}^\mathcal{A}$ is the tensor of decoder hidden states of all agents in the scene.
This update allows us to ground the tokens to the relevant agents' states so that the decoded trajectories can respond to the change of other agents.

\paragraph{Token Attention}
To determine which tokens are relevant to the trajectory at the current time step, we consider the attention structure that has been used widely in machine translation~\cite{luong2015} to generate the attention over the input tokens for the decoder.
We re-encode the word embeddings at every time step using an LSTM token encoder since the word embeddings are updated with relevant agents' state.
The token attention module takes the last output $c_t$ from the LSTM token encoder along with the current agent's trajectory decoder state in the previous time step $h_{t-1}$ to generate attention over tokens as in \cite{luong2015}.
This attention is used to pool the word embeddings of the generated language tokens to provide the context vector $c_t^\prime$ to decode the trajectory:
\begin{equation}
    c_t^\prime = \textit{Attention}(c_t, h_{t-1}) \cdot \rev{Y^\prime_{1:M}}
\end{equation}
Finally, this context vector is concatenated with the agent's state to decode the trajectory $s_t$ and update the agent's decoder state $h_t$ at time $t$ using an LSTM.

\subsection{Training the Language-based Predictor}
We train the generator and the discriminator end-to-end.
When generating trajectories $\hat{\mathbf{s}}_F$, similar to \cite{gupta2018social}, we compute the Minimum over N (MoN) losses using the average displacement error (ADE) to encourage the model to cover the ground-truth positions:
\begin{equation}
    \mathcal{L}_{\textit{MoN}} = \min_n(\textit{ADE}(\hat{\mathbf{s}}_F)^{(n)})
\end{equation}
When the ground-truth description is available, we augment the trajectory generation loss by a cross entropy loss for the generated descriptions:
\begin{equation}
    \mathcal{L}_{\textit{lang}} = \frac{1}{M} \sum_{m=0}^M \textit{CrossEntropy}(y_m, \hat{y}_m)
\end{equation}
The total loss for the generator is a combination of the losses above:
\begin{equation}
    \mathcal{L}_G = \lambda_1 \mathcal{L}_{MoN} + \lambda_2 \mathcal{L}_{\textit{lang}}
\end{equation}
Note that this loss structure affords training of the generator even when the ground-truth descriptions do not cover all trajectories, and merely uses descriptions as an additional supervision.
This is especially true for multiple agents -- all agents share the same latent structure, but only one agent has a ground-truth description loss associated with it.
The parameter sharing ensures that other agents are also explainable.
Finally, the discriminator uses a binary cross entropy loss to classify the generated trajectories and the ground-truth trajectories.

\section{EXPERIMENTS}

To train and evaluate the model with linguistic representations, we describe the procedure to augment the existing datasets for training. Then we describe the baseline models to test the ability to predict future trajectories and demonstrate the scenarios where language can improve predictions.

\subsection{Dataset}

\begin{figure}
\centering
\scalebox{0.8}{
\begin{tabular}{lc}
Time horizon & 2s past, 3s future \\
Total \# of trajectories & 7,753,060 \\
\# of trajectories with descriptions &  1,168,963 \\
\# of descriptions referring to other agents & 272,877 \\
Average number of tokens per description & 3.17 \\
Vocabulary size & 16 \\
\end{tabular}
}
\caption{The overall statistics of the dataset augmented with language descriptions. We do not augment all trajectories with captions because some trajectories are too short or the agent does not move.}
\label{table:data-stats}
\end{figure}

There is no existing motion prediction dataset that contains annotated language descriptions.
To train a predictor that utilizes the language as a latent representation, we first produce a dataset that contains trajectory and description pairs.
We augment the Argoverse v1.1~\cite{chang2019argoverse} with synthetic language descriptions.
The statistics of the training dataset are summarized in Table~\ref{table:data-stats}.
We consider trajectories from all agents in a scenario as candidates for annotation with linguistic descriptions, not just the trajectories from the ego vehicles.

We generate the description for each trajectory by applying a set of predefined filters to identify meaningful parts of trajectories.
Augmenting with synthetic descriptions gives us higher language coverage in trajectories than is feasible with human annotations.
However, the token and structure are bounded by the types of filters we employ.

We generate tokens \texttt{MoveFast}, \texttt{MoveSlow}, \texttt{Stop}, \texttt{TurnLeft}, \texttt{TurnRight}, \texttt{SpeedUp}, \texttt{SlowDown} based on the velocity, angular velocity, and acceleration of trajectories.
We generate tokens for lane changes (\texttt{LaneKeep}, \texttt{LaneChangeLeft}, \texttt{LaneChangeRight}) based on the change of the closest lane centerlines.
Two tokens involving other agents are generated: \texttt{Follow Agent\#k} and \texttt{Yield Agent\#k}, based on intersection tests and overlaps.
%
%
In the case of \texttt{Yield}, the target agent needs to slow down and arrive at the intersecting area later than the agent it yields to.

The tokens are organized in a sequence according to their temporal order.
If there are multiple tokens active at the same time, we randomly select one of them, to imitate how humans selectively describe some properties of trajectories instead of providing every minute detail.
%
%
We do not label trajectories if they are too short or if the generated tokens are oscillating between tokens with opposite meanings.

\subsection{Experiment Setup}

\subsubsection{Baselines}
As described in the related work section, no existing model is trained with language descriptions or uses language as intermediate representations for trajectory predictions.
We compare our model to baselines and perform two ablations.
These baseline models all use the same encoder as described here but implement different decoders.
\rev{The first baseline is a vanilla LSTM decoder without employing any structure, modules, or attention mechanism}.
The second baseline is an LSTM decoder with multi-head attention~\cite{mercat2020multi}.
These attention heads attend to other agents' LSTM states at time $t-1$.
This represents the predictors that model the interactions between agents explicitly and is the model closest to ours, with standard methods of capturing social interactions \cite{gupta2018social,amirian2019social}, map information \cite{gao2020vectornet}, and attention structures \cite{mercat2020multi}.

Additionally, we consider two ablations.
\textit{Ours (no attention)} is our model without both attention modules in the predictor.
The decoder LSTM directly concatenates the last output from the token encoder to generate trajectories.
This shows the performance of language embeddings but doesn't consider how language relates to trajectories.
\textit{Ours (no agent attention)} ablates the agent's attention but keeps the token attention.
This ablation doesn't update the token that refers to other agents with the pooled agent states. 
This shows the performance when only the temporal aspects of language are considered.

\subsubsection{Model Details}
The encoder LSTM has a hidden dimension size of 32 and output dimension of 32.
In the decoder, we use a dimension of 4 for the token generator LSTM hidden state and the token embedding.
The attention modules are one-layer MLPs with hidden dimension 4.
We co-train the word embeddings with the rest of the network.
We use 0.2 dropout for the token generation, encoder, and attention modules.
For the multi-head attention baseline, we follow \citet{mercat2020multi} in using six 10-dimensional attention heads.
The loss coefficients are selected to be 1.
The model is optimized using Adam and trained on an NVIDIA Tesla V100 GPU with learning rate 1e-3 and batch size 32.
At every training epoch, we rebalance the training examples that contain and don't contain linguistic descriptions to make sure the model receives smooth gradients from the language data.

\subsubsection{Metrics}
We measure the correctness of the predicted language by computing its recall, i.e. the fraction of ground-truth tokens which are predicted.
We choose recall because the additional predicted tokens may not be wrong, but are just missing in the ground-truth (\rev{the predefined filters} might not have \rev{identified the maneuvers or interactions}).
The prediction performance of trajectories is evaluated with minimum average displacement error (ADE) and final displacement error (FDE)~\cite{gupta2018social} in meters.
The ADE measures the average distance between the predicted and the ground-truth trajectories.
The FDE measures the final distance at the selected prediction horizons.
We use minimum-of-N (MoN, variety loss) with $N=6$ unless stated otherwise.

In order to compare how language improves trajectories at long time horizons, we
report displacement errors at 1 and 3 seconds.
We produce results with the number of samples, $N$, set to 6.
We also compute the entropy of the predicted trajectory samples to understand how language influence the trajectory choices to more relevant ones.
Similar to Trajectron++~\cite{salzmann2020trajectron++}, we perform Gaussian kernel density estimation on the samples to compute the entropy of the samples.
For baseline models, the entropy is $H(\mathbf{s}_F)$; for the language-based models, this is the conditional entropy $H(\mathbf{s}_F|Y_{1:M})$.

\subsection{Results}
\subsubsection{Quantitative Results}

\begin{figure}
    \centering
    \scalebox{0.8}{
    \begin{tabular}{lccc}
      Model & ADE & FDE & Entropy \\
      \hline \hline
      LSTM decoder & 0.69 & 1.46 & 2.66 \\
      Multihead attention decoder  & 0.67 & 1.39 & 2.51 \\
      Ours (no attention) & 0.67 & 1.44 & 2.61 \\
      Ours (no agent attention) & 0.67 & 1.39 & 2.56 \\
      Ours & 0.65 & 1.35 & 2.41 \\
    \end{tabular}}
    \caption{MoN average displacement errors (ADE) and final displacement errors (FDE) of our method and baseline models with $N$=6 samples on the Argoverse dataset predicting 3 seconds in the future.}
    \label{tab:argo-mon}
\end{figure}


\begin{figure}
    \centering
    \scalebox{0.75}{
    \begin{tikzpicture}
      \pgfplotsset{every tick label/.append style={font=\footnotesize}}
      \begin{axis} [ybar,bar width=3pt,
        ymin=0,ymax=0.4,
        height=5cm,width=9cm,
        xticklabels={MoveFast,MoveSlow,SpeedUp,SlowDown,TurnLeft,TurnRight,LaneChangeLeft,LaneChangeRight,LaneKeep,Agent \#k,Follow,Yield},
        xtick={1,...,12},
        x tick label style={rotate=30,anchor=east}]
        \addplot coordinates {
          (1,0.375) (2,0.356) (3,0.001) (4,0.012) (5,0.013) (6,0.008)
          (7,0.004) (8,0.003) (9,0.213) (10,0.008) (11,0.007) (12,0.001)
        };
        \addplot coordinates {
          (1,0.212) (2,0.215) (3,0.035) (4,0.116) (5,0.095) (6,0.045)
          (7,0.020) (8,0.019) (9,0.145) (10,0.077) (11,0.018) (12,0.001)
        };
        \legend {Ground-truth, Predicted};
      \end{axis}
    \end{tikzpicture}}
    \vspace{-2ex}
    \caption{Distributions of ground-truth and generated tokens in the validation set. \rev{In our experiment, k=\{1,2,3,4\} indicating the sentence can refer to up to four agents}. Note that we generate language tokens for all trajectories but only have ground-truth for the confident examples, resulting in gaps between the predicted and ground-truth distributions.}
    \label{fig:token_dist}
\end{figure}
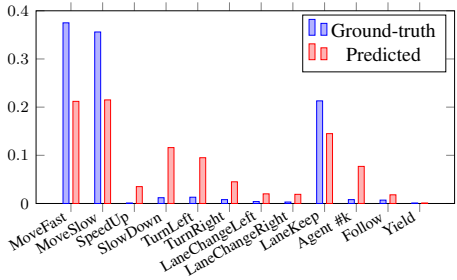

The recall of the generated linguistic descriptions is 85.7\%.
\cref{fig:token_dist} shows the distribution of ground-truth tokens along with the generated tokens in the validation set.
Even though the ground-truth tokens mostly cover simple maneuvers such as speed, our training regime identify the trajectories with rarer tokens while mimicking the ground-truth distributions.

\cref{tab:argo-mon} summarizes the displacement errors and entropy for Argoverse at 3 seconds.
Our model produces the lowest FDE and ADE compared to baselines.
Ablation of any attention module shows poor performance and higher entropy as the trajectory decoder does not know what the encoded token vectors mean.

We also computed the FDE and ADE at 1 second.
All baseline models have 0.29 FDE and 0.40 ADE.
Our model has 0.28 FDE and 0.40 ADE.
Comparing the displacement errors at 1 vs. 3 seconds confirms that the influence of language is more prominent at longer horizons.
\rev{We observe the reduction of entropy compared to the baselines. In order to gauge if this reduction is from the information stored in the generated sentences,} we compute the information gain based on the entropy of predictions conditioned on the generated descriptions and all-padded descriptions.
%
%
%
The average information gain for language conditioned predictions is 0.48 bits.
This is a significant influence as many trajectories in Argoverse have one or two tokens in the description.
Furthermore, we did not have the linguistic representation bottlenecked \cite{chechik2005information,chen2016infogan} in the predictor, which may lead to underestimating the mutual information. 

%
%

%

\subsubsection{Qualitative Results}

\begin{figure}[htp]
\centering
\scalebox{0.8}{
\begin{tabular}{p{15ex}c}
Trajectories & Token attention at $t$ \\
\hline
    \includegraphics[width=0.13\textwidth]{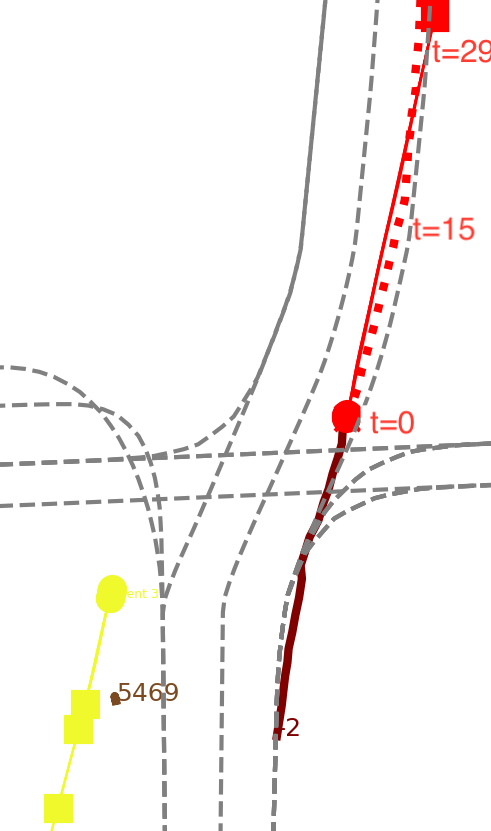}
    \scalebox{0.85}{
    \begin{minipage} [t] {0.3\textwidth}
    \begin{itemize}
        \item Ground truth: MoveSlow
        \item Predicted: MoveSlow, LaneKeep
    \end{itemize}
    \end{minipage}}
    &
    \raisebox{22ex}{
    \begin{minipage} [t] {0.3\textwidth}
    \begin{itemize}
        \item $t=0$: \colorbox{red!61}{MoveSlow}  \colorbox{red!39}{LaneKeep}\\
        \item $t=15$: \colorbox{red!50}{MoveSlow}  \colorbox{red!50}{LaneKeep}\\
        \item $t=29$: \colorbox{red!28}{MoveSlow}  \colorbox{red!72}{LaneKeep}
    \end{itemize}
    \end{minipage}}
    \\
    & \\
    \hline
    \includegraphics[width=0.06\textwidth]{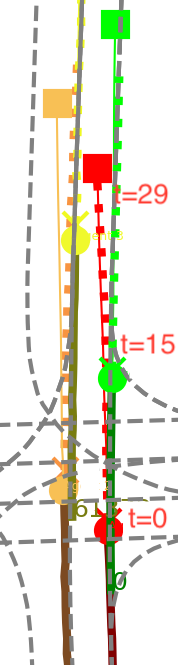}
    \scalebox{0.85}{
    \begin{minipage} [t] {0.35\textwidth}
    \begin{itemize}
        \item Ground truth: LaneChangeLeft
        \item Predicted: MoveSlow, LaneChangeLeft
    \end{itemize}
    \end{minipage}} &
    \raisebox{22ex}{
    \begin{minipage} [t] {0.35\textwidth}
    \begin{itemize}
        \item $t=0$: \colorbox{red!81}{MoveSlow}  \colorbox{red!29}{LaneChangeLeft}\\
        \item $t=15$: \colorbox{red!30}{MoveSlow}  \colorbox{red!70}{LaneChangeLeft}\\
        \item $t=29$: \colorbox{red!30}{MoveSlow}  \colorbox{red!70}{LaneChangeLeft}
    \end{itemize}
    \end{minipage}}\\
\end{tabular}}
\caption{Example predicted trajectories and linguistic descriptions. Red lines on the left correspond to the target agent. The dark red line is the observed past trajectory. The solid red line is the predicted future sample and the dotted red line is the ground-truth trajectory. We also show the attention weights of the generated language tokens at different time steps on the right. Tokens with darker background are the ones attended to more when generating trajectory points at that time step. The changes in the background color, i.e. in attention, indicates a transition between maneuvers.}
\label{fig:example}
\end{figure}

In \cref{fig:example}, we present qualitative examples that demonstrate the effectiveness of linguistic representations to predict trajectories.
We visualize the attention weights for the predicted tokens at different time steps (start, middle, and end of the trajectory) to show which tokens are in the focus as trajectories are generated.
This visualization also shows the utility of attention for understanding the inferences of the model.
The generated descriptions may not match all the ground-truth tokens but that is not necessarily a prediction error.
For example, the first row in \cref{fig:example} only has \texttt{MoveSlow} as ground truth while the predicted tokens include \texttt{LaneKeep}.
This is because we only include the most confident tokens in the description and sample only one token if tokens overlap in time.
The model is able to recover missing tokens during training.

\begin{figure}[htp]
    \centering
    \scalebox{0.8}{
    \begin{tabular}{cp{2ex}c}
      \includegraphics[width=0.2\linewidth]{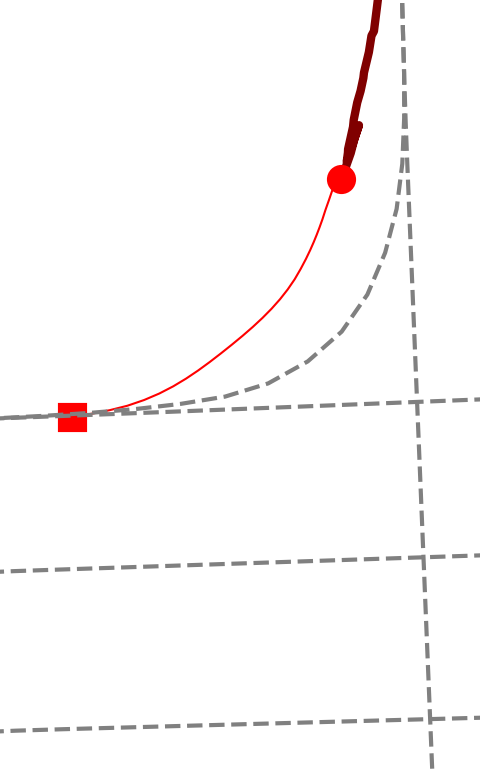} & &
      \includegraphics[width=0.2\linewidth]{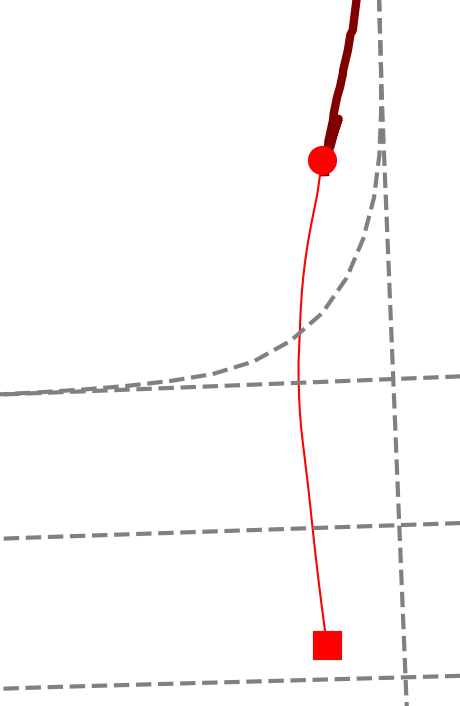}\\
      \small{(a) turn right} & & \small{(b) turn left} \\
    \end{tabular}}
    \caption{The trajectory changes after changing a token in the linguistic description: turn right vs. turn left. The dark red line is the observed past trajectory and the thin red line is the generated future trajectories.}
    \label{fig:diff_dir}
\end{figure}

We can also directly modify the generated descriptions to see how changing this internal reasoning affects the predicted trajectories.
This allows us to reveal the meaning of different tokens.
In \cref{fig:diff_dir}, we demonstrate that when we observe the same past trajectory but change the tokens from ``turn left'' to ``turn right'', the predicted trajectories of the agent change accordingly.
%

\begin{figure}
    \centering
    \begin{tabular}{@{}c@{}c@{}}
        \includegraphics[width=0.065\textwidth]{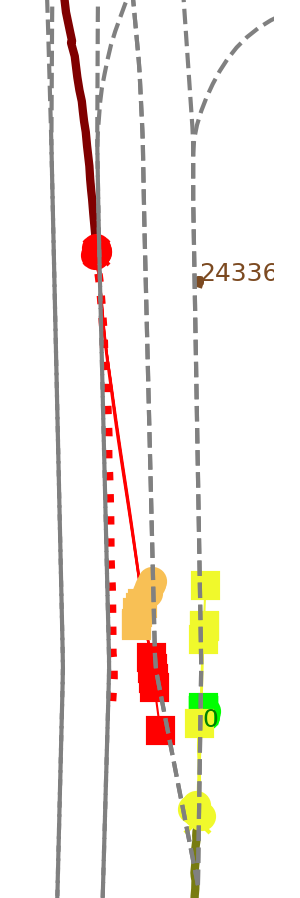}
        &
        \includegraphics[width=0.23\textwidth]{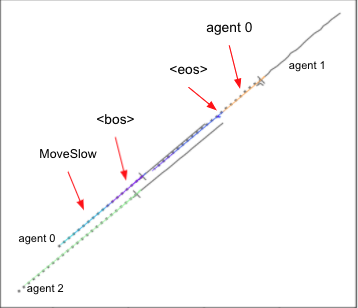}
        \\
        \scalebox{0.55}{
        \begin{minipage} [t] {0.25\textwidth}
            (a) Example failure\\[-3ex]
            \begin{itemize}
            \item Ground truth: LaneKeep
            \item Predicted: TurnLeft
        \end{itemize}
        \end{minipage}}
        &
        \scalebox{0.55}{
        \begin{minipage} [t] {0.5\textwidth}
          (b) Example token association: The paths for agent 0 and 1 are colored and marked with the most relevant tokens showing that agent 0 slowed down so agent 1 needed to attend to it.
        \end{minipage}}
        \\
    \end{tabular}

    \caption{Example of introspection using language. (a) The error in the predicted trajectories is reflected in the generated sentence. The dotted red line is the ground truth. (b) The most relevant tokens at different part of trajectories.}
    \label{fig:understand_traj}
\end{figure}

This representation also provides a mechanism to introspect the predictor's failures; see \cref{fig:understand_traj} for example predicted tokens and trajectories in this case.
When there is a prediction error, this is evident from the predicted description.
We can also find the relevant parts of trajectories that correspond to different
errors by inspecting the attention weights on different tokens; see
\cref{fig:understand_traj}(b) for an example of interpreting interactions
between two agents.
In the example in \cref{fig:understand_traj}(a), the cause of the error can be identified from the generated description \emph{TurnLeft}.
Practically, this can enable one to collect more targeted training examples.
An end-to-end model without an intermediate representation that can be inspected
and understood by humans would not provide such clear guidance about what went
awry and how to fix it.

\section{EXTENDING TO NATURAL LANGUAGE}

While the predictors shown thus far are based on a synthetic language, \rev{we want to understand if it is feasible to train the model with natural language}.
We sampled 40,000 trajectories from the Waymo dataset~\cite{ettinger2021waymo} and had humans annotate them with captions.
We observed that the annotated sentences use a much more diverse vocabulary than the synthetic language.
For example, humans use ``drive after'' or ``drive behind'' to describe following another agent.
These captions also contain temporal and spatial relationships such as ``before'', ``left'', and ``right''.
When processing natural language, we employ an English tokenizer to preprocess the human-generated captions and then embed the predicted tokens using GloVe~\cite{pennington2014glove}.
The rest of the network is unchanged aside from using a hidden dimension of 16 for the language generator and encoder modules because the space of possible words, i.e. tokens, has increased dramatically.
This extension generates naturalistic sentences for the predictor as in \cref{fig:waymo_example}.

\begin{figure}
    \centering
    \scalebox{0.75}{
    \begin{tabular}{l|l}
        The agent starts to slow down. &
        The agent slows down behind an agent. \\
        The agent turns behind an agent. &
        The agent slows down because light. \\
        The agent crosses intersection fast. &
        The agent slows down waiting an agent. \\
        The agent halts on crosswalk. &
        The agent moves fast passing an agent. \\
        The agent passing bumps slow. &
        The agent takes left while crossing. \\
    \end{tabular}}
    \caption{Example of predicted sentences in the Waymo dataset.}
    \label{fig:waymo_example}
\end{figure}

\section{CONCLUSION \& FUTURE WORK}
We propose a new trajectory prediction model that employs language as an intermediate representation. 
Our method generates interpretable linguistic tokens and is sensitive to these tokens.
The attention weights associated with each token is meaningful and can be used to understand what each part of a linguistic description is referring to.
While we don't explore this here, the learned language model can provide an interface by which humans can include their preferences into the reasoning of the car.
%
%
Such interfaces can also provide a level of comfort to users by creating meaningful explanations for the behavior of the car.
%

Our proposed model is based on single agent's observations and view.
For predicting multiple agents in a scene, disagreements between the linguistic descriptions made by different agents, the Rashomon effect~\cite{heider1988rashomon}, may arise when predicting multiagent interactions.
Contrasting our linguistic descriptions with whole-scene multimodal language-vision representations could be an interesting step forward.



\section*{\small{ACKNOWLEDGMENT}}
\small{
This work was supported by the Center for Brains, Minds and Machines, NSF STC award 1231216, 
the MIT CSAIL Systems that Learn Initiative, the CBMM-Siemens Graduate Fellowship,
the DARPA Artificial Social Intelligence for Successful Teams (ASIST) program,
the United States Air Force Research Laboratory 
and United States Air Force Artificial Intelligence Accelerator under Cooperative Agreement Number FA8750-19-2-1000, and the Office of Naval Research under Award Number N00014-20-1-2589 and Award Number N00014-20-1-2643.
The views and conclusions contained in this document are those of the authors and should not be interpreted as representing the official policies, either expressed or implied, of the U.S. Government. The U.S. Government is authorized to reproduce and distribute reprints for Government purposes notwithstanding any copyright notation herein.
}

\bibliographystyle{IEEEtranN}
\renewcommand*{\bibfont}{\footnotesize}
\bibliography{reference}

\end{document}